\documentclass[10pt,twocolumn,letterpaper]{article}

\usepackage{cvpr}
\usepackage{times}
\usepackage{epsfig}
\usepackage{graphicx}
\usepackage{amsmath}
\usepackage{amssymb}

\usepackage{url}
\usepackage[caption=false]{subfig}
\usepackage{mdwlist}
\usepackage[lined, boxed]{algorithm2e}
\usepackage{multirow}
\usepackage{amsthm}
\usepackage{color}

\graphicspath{{figs/}}

\newtheorem{theorem}{Theorem}

\usepackage[pagebackref=true,breaklinks=true,letterpaper=true,colorlinks,bookmarks=false]{hyperref}

\cvprfinalcopy 

\ifcvprfinal\fi
\begin{document}

\title{Not Afraid of the Dark: NIR-VIS Face Recognition via Cross-spectral Hallucination and Low-rank Embedding}

\author{Jos\'e Lezama$^1$\thanks{Denotes equal contribution.}, Qiang Qiu$^{2*}$  and Guillermo Sapiro$^2$\\
$^1$IIE, Universidad de la Rep\'ublica, Uruguay. $^2$ECE, Duke University, USA. 
}

\maketitle

\begin{abstract}
Surveillance cameras today often capture NIR (near infrared) images in low-light
environments. However, most face datasets accessible for training and
verification are only collected in the VIS (visible light) spectrum. It remains
a challenging problem to match NIR to VIS face images due to the different light
spectrum.  Recently, breakthroughs have been made for VIS face recognition by
applying deep learning on a huge amount of labeled VIS face samples. The same
deep learning approach cannot be simply applied to NIR face recognition for two
main reasons: First, much limited NIR face images are available for training
compared to the VIS spectrum. Second, face galleries to be matched are mostly
available only in the VIS spectrum.  In this paper, we propose an approach to
extend the deep learning breakthrough for VIS face recognition to the NIR
spectrum, without retraining the underlying deep models that see only VIS
faces. Our approach consists of two core components, cross-spectral
hallucination and low-rank embedding, to optimize respectively input and output
of a VIS deep model for cross-spectral face recognition. Cross-spectral
hallucination produces VIS faces from NIR images through a deep learning
approach. Low-rank embedding restores a low-rank structure for faces deep
features across both NIR and VIS spectrum.  We observe that it is often equally
effective to perform hallucination to input NIR images or low-rank embedding to
output deep features for a VIS deep model for cross-spectral recognition. When
hallucination and low-rank embedding are deployed together, we observe
significant further improvement; we obtain state-of-the-art accuracy on the
CASIA NIR-VIS v2.0 benchmark, without the need at all to re-train the
recognition system.
\end{abstract}

\begin{figure} [t]
\centering
\includegraphics[angle=0, width=.49\textwidth]{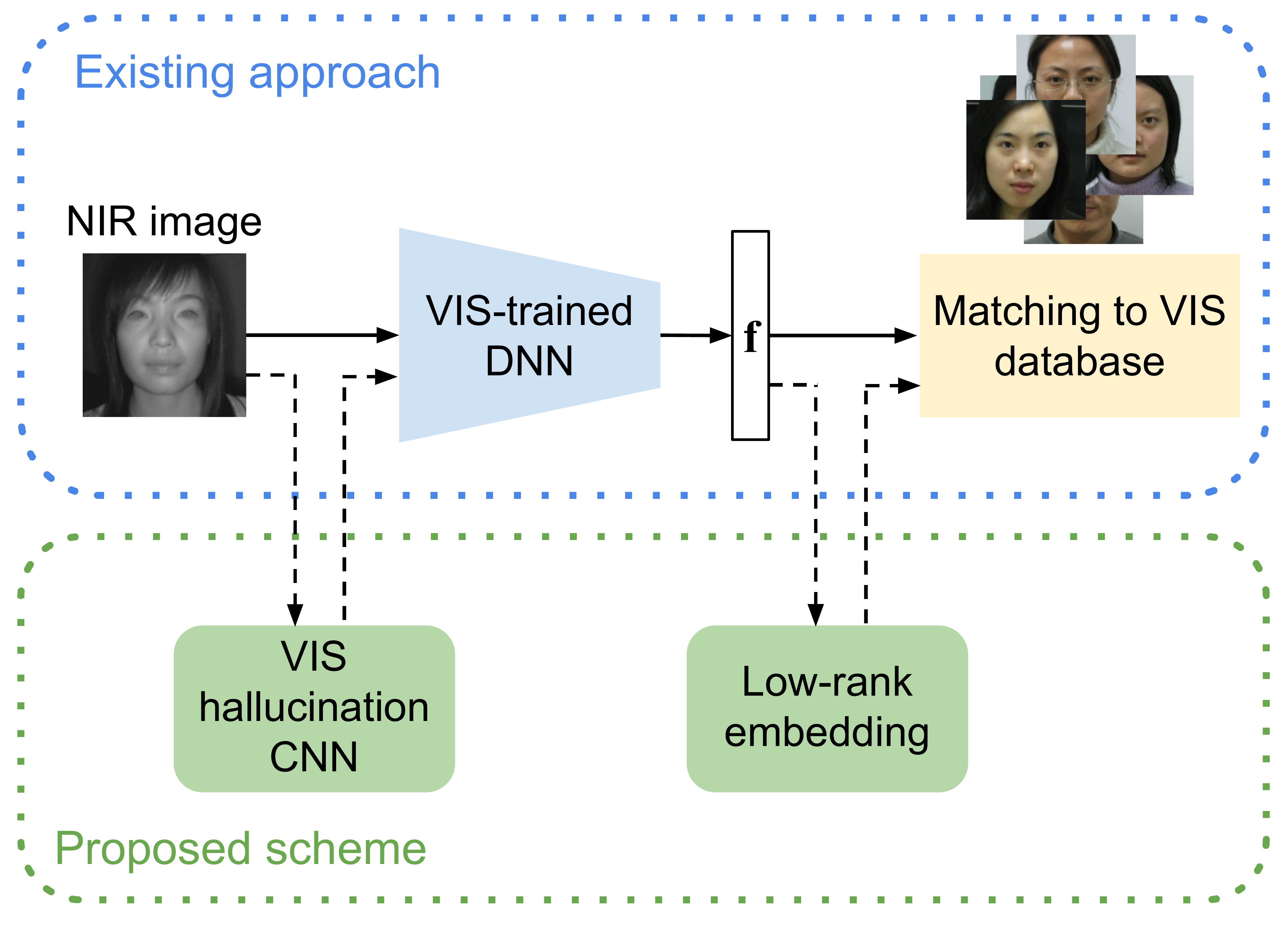}
\caption{Diagram of the proposed approach. A simple NIR-VIS face recognition
  system consists in using a Deep Neural Network (DNN) trained only on VIS
  images to extract a feature vector \textbf{f} from a NIR image and use it for
  matching to a VIS database. We propose two modifications to this basic
  system. First, we modify the input by hallucinating a VIS image from the NIR
  sample. Secondly, we apply a low-rank embedding of the DNN features at the
  output. Each of this modifications produces important improvements in the
  recognition performance, and an even greater one when applied together.}
\label{fig:system_diagram}
\end{figure}

\section{Introduction}

In a typical forensic application involving night-time surveillance
cameras, a probe image of an individual is captured in the
near-infrared spectrum (NIR), and the individual must be recognized out of a
gallery of visible spectrum (VIS) images.  Whilst VIS 
face recognition is an extensively studied subject, face
recognition in the NIR spectrum remains a relatively unexplored field.

In standard VIS face recognition, impressive progress has been achieved
recently. This is due in part to the excellent performance of deep neural
networks \cite{vggface, schroff2015facenet, taigman2014deepface,
  yi2014learning}, which benefit from the availability of very large datasets of
face photos, typically mined from the Internet \cite{huang2007labeled,
  yi2014learning}. Such a fruitful strategy for data collection cannot be
applied to NIR images, where training data is much scarcer.

Naturally, one would like to leverage the power of state-of-the-art
VIS face recognition methods and apply them to NIR. Recent works have
made significant progress in this direction
\cite{juefei2015nir,saxena2016heterogeneous}, but the recognition
rates remain much lower than the ones achieved in the VIS spectrum. In
this work, we take a major step towards closing that gap.

We consider using a pre-trained deep neural network (DNN) which has
seen only VIS images as a black-box feature extractor, and propose
convenient processing of the input and output of the DNN that produce
a significant gain in NIR-VIS recognition performance.  The proposed
approach, summarized in Figure~\ref{fig:system_diagram}, consists of
two components, cross-spectral hallucination and low-rank embedding,
to optimize respectively the input and output of a pre-trained VIS DNN
model for cross-spectral face recognition.

First, we propose to modify the NIR probe images using deep cross-spectral
hallucination based on a convolutional neural network (CNN).\footnote{To avoid
  confusion, in this paper we will refer to the deep neural network used for
  feature extraction as DNN and to the convolutional neural network used to
  hallucinate full-resolution VIS images from NIR as CNN.} The CNN learns a
NIR-VIS mapping on a patch-to-patch basis. Then, instead of inputting the NIR
probe directly to the feature extraction DNN, we input the cross-spectrally
hallucinated. Secondly, we propose to embed the output features of the DNN using
a convenient low-rank transform \cite{lrt}, making the transformed NIR and VIS
features of one subject lie in the same low-dimensional space, while separating
them from the other subjects' representations. While here illustrated for the
important problem of face recognition, this work provides a potential new
direction that combines transfer learning (at the input/output) with joint
embedding (at the output).

The two proposed strategies achieve state-of-the-art results
when applied separately, and achieve an even more significant
improvement when applied in combination. We demonstrate that
both techniques work well independently of the DNN used for feature
extraction.

\section{Related Work}
One common strategy for NIR-VIS face recognition, employed since the early work
of Yi et al. \cite{yi2007face}, is to find a mapping of both the NIR and VIS
features to a shared subspace, where matching and retrieval can be performed for
both spectrums at the same time. This metric learning strategy was applied
in many successive works
\cite{fernando2013unsupervised,hou2014domain,lei2009coupled,mignon2012cmml}, and
more recently using DNN-extracted features
\cite{vggface,saxena2016heterogeneous}. 
Most metric learning methods learn metrics with triplet \cite{lmnn} or pairwise \cite{itml} constraints.
The triplet loss is often adopted in deep face models to learn a face embedding \cite{vggface, schroff2015facenet}.
Saxena and Verbeek
\cite{saxena2016heterogeneous} studied the performance implications of using
different feature layers of the DNN in combination with different metric
learning approaches, and propose a combination of the two. In this work, we
consider the well developed DNN as a black box and use the features produced by
the DNN at the second-to-last layer. We adopt a low-rank constraint
 to learn a face embedding, which proves effective for cross-spectral tasks.

Another strategy is to convert the NIR probe to a VIS image
\cite{juefei2015nir,li2008hallucinating,sarfraz2015deep} and apply standard VIS
face recognition to the converted version of the probe. One of the first to
utilize this strategy for face VIS hallucination, Li et
al. \cite{li2008hallucinating}, learn a patch-based linear mapping from a
middle- and long-wavelength infrared (MW-/LWIR) image to a VIS image, and
regularize the resulting patches with an MRF. Juefei et
al. \cite{juefei2015nir}, used a cross-spectral dictionary learning approach to
successfully map a NIR face image to the VIS spectrum. On the obtained VIS
image, they apply Local Binary Patterns (LBP) \cite{ahonen2004face} and a linear
classifier to perform face recognition. Converting from infrared to VIS is a
very challenging problem, but has the clear advantage of allowing to use
existing traditional face recognition algorithms on the converted images. To the
best of our knowledge, this is the first time a deep learning approach is used
to hallucinate VIS faces from NIR.

Several works exist for the task of cross-spectral conversion of outdoor scenes
\cite{hogervorst2010fast,toet2012progress,zheng2008local}. This scenario has the
advantage that more multispectral data exists for generic scenes
\cite{BS11}. Building a dataset of cross-spectral face imaging with 
correct alignment for a significant number of subjects is a much more challenging
task. We believe it is in part due to this difficulty that few works exist in this
direction.

Given the advantage of thermal images not requiring a light source, a lot of
attention has been given to the thermal to VIS face recognition task
\cite{bourlai2012multi, cao2014matching, sarfraz2015deep}. Related to our work,
Sarfraz et al. \cite{sarfraz2015deep} used a neural network to learn the reverse
mapping, from VIS to MW-/LWIR, so that a thermal face image could be matched to
its VIS counterpart. This strategy has the disadvantage of having to apply the
mapping to each VIS image in the dataset. We propose to use a convolutional neural
network to compute the mapping between the (single) tested NIR and VIS images.

Another important family of work on cross-spectral face recognition focuses on the
features used for recognition, and suggested strategies include engineering light
source invariant features (LSIF) \cite{liu2012heterogeneous}, performing
cross-domain feature learning \cite{Jin_etal,lu_etal,Yi_etal,zhu2014matching},
and adapting traditional hand-crafted features \cite{dhamecha2014effectiveness,
  klare2010heterogeneous}.

Alternative approaches fit existing deep neural networks to a given database,
e.g., \cite{liu2016transferring}, achieving as expected improvements in the
results for that particular dataset. Contrary to those, the generalization power
of our proposed framework is born from the technique itself; without any kind of
re-training we achieve state-of-the-art results. This is obtained while enjoying
the hard work (and huge training) done for existing networks, simply by adding
trivial hallucination and linear steps. As underlying networks improve, the
proposed framework here introduced will potentially continue to improve without
expenses on training or data collection.

In this paper, we build on the ideas of \cite{saxena2016heterogeneous} and
\cite{juefei2015nir}. We use a DNN pre-trained on a huge dataset of VIS images
as a feature extractor. At the input of this DNN, we propose to preprocess the
NIR input using deep cross-modal hallucination.  At the output, we propose to
embed the feature vector using a low-rank transform. The simplicity of the
approach and the use of off-the-shelf well optimized algorithms is part of the
virtue of this work. As a side contribution, we derive a secondary dataset from
the CASIA NIR-VIS 2.0 dataset \cite{NIR-VIS-data} consisting of more than 1.2
million pairs of aligned NIR-VIS patches.

\section{Cross-spectral Hallucination}\label{sec:hallucination}
Most if not all DNN face models are designed and trained to operate on VIS face
images, thanks to the availability of enormous VIS face datasets. It is to
expect that such deep models do not achieve their full potential when their
input is a NIR face image. In this section we propose to preprocess the NIR
image using a convolutional neural network (CNN) that performs a cross-spectral
conversion of the NIR image into the VIS spectrum. We will show that using the
hallucinated VIS image as input to the feature extraction DNN, instead of the
raw NIR, produces a significant gain in the recognition performance. Note that
the goal here is not necessarily to produce a visually pleasant image, but one
that better fits the NIR-VIS DNN based matching.

The cross-spectral hallucination CNN is trained on pairs of corresponding
NIR-VIS patches that are mined from a publicly available dataset, as will be
described below. In the VIS domain, we work in a luminance-chroma color space,
since it concentrates the important image details in the luminance channel and
minimizes the correlation between channels, making the learning more
efficient. We find the YCbCr space to give the best results, and we observe no
difference between training the three channels with shared layers or
independently. For simplification we train three different networks. Because the
luminance channel Y contains most of the subject's information, we utilize a
bigger network for this channel and smaller networks for the two chromas. Also,
because the blue component varies very little in faces, an even smaller network
is enough for the blue-difference chroma Cb. The architecture of the networks is
shown in Table~\ref{tab:cross-spectral-cnn-architecture}.  The three networks
receive a 40x40 input NIR patch and consist of successive convolutional layers
with PReLU activation functions (except in the last layer) \cite{he2015delving},
and an Euclidean loss function at the end. We pad each layer with zeros to have
the same size at the input and output. The three networks have an hour-glass
structure where the depth of the middle layers is narrower than the first and
last layers. This makes for efficient training while allowing highly non-linear
relations to be learned.

\begin{figure} [ht]
\centering
\renewcommand{\arraystretch}{.4}
\begin{tabular}{
    c@{\hspace{.2em}}
    c@{\hspace{.2em}}
    c@{\hspace{.2em}}
    c@{\hspace{.2em}}}
\includegraphics[angle=0, width=0.105\textwidth]{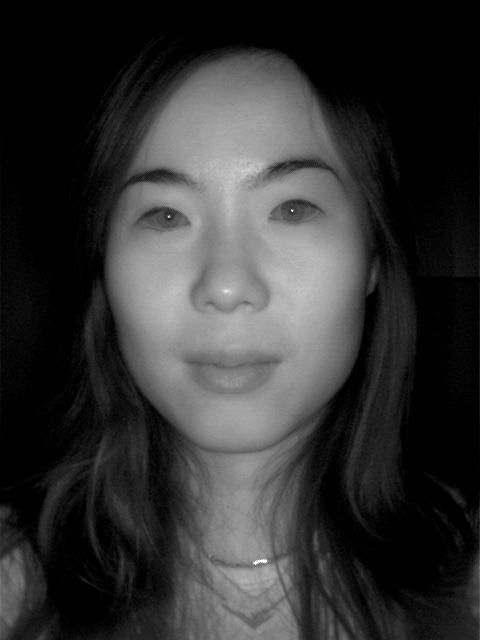} &
\includegraphics[angle=0, width=0.105\textwidth]{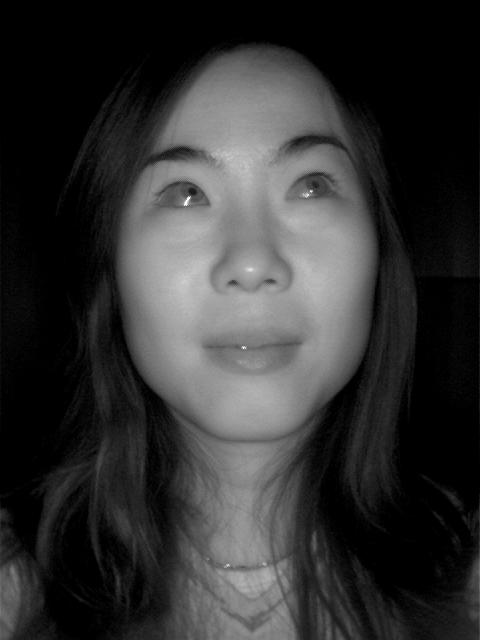} &
\includegraphics[angle=0, width=0.105\textwidth]{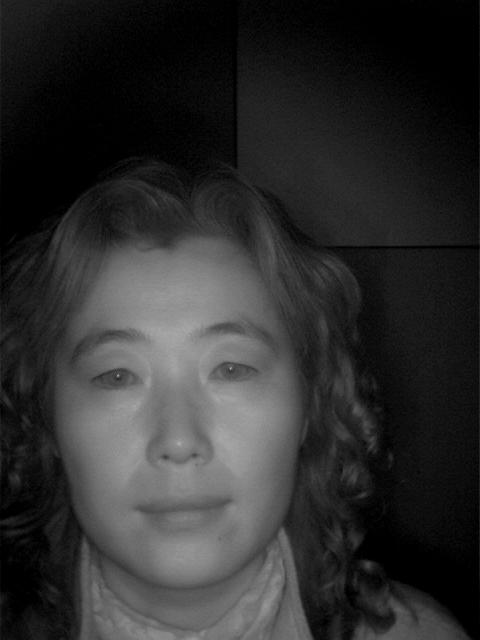} &
\includegraphics[angle=0, width=0.105\textwidth]{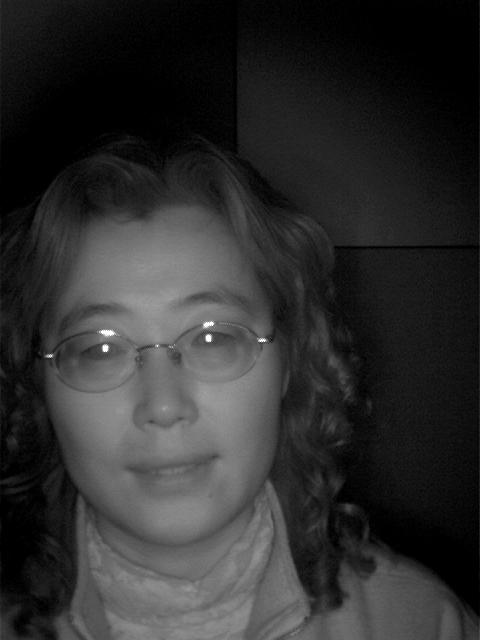} \\
\includegraphics[angle=0, width=0.105\textwidth]{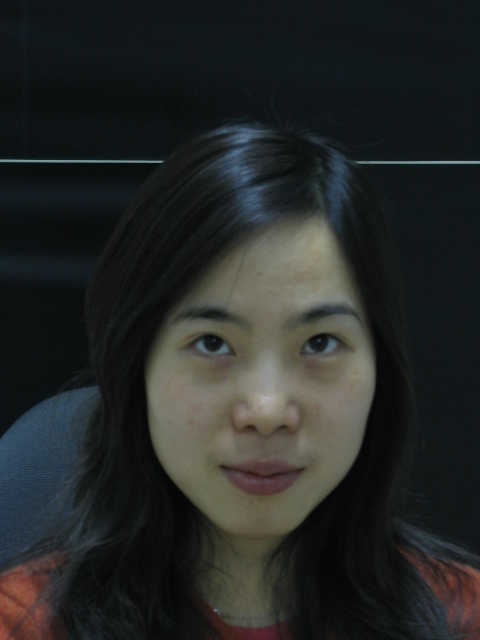} &
\includegraphics[angle=0, width=0.105\textwidth]{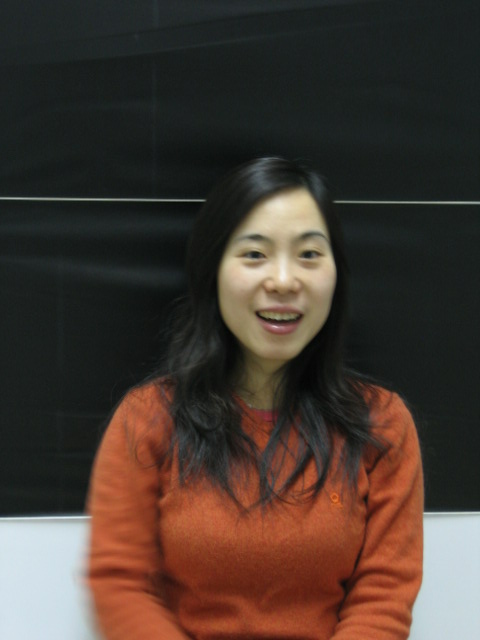} &
\includegraphics[angle=0, width=0.105\textwidth]{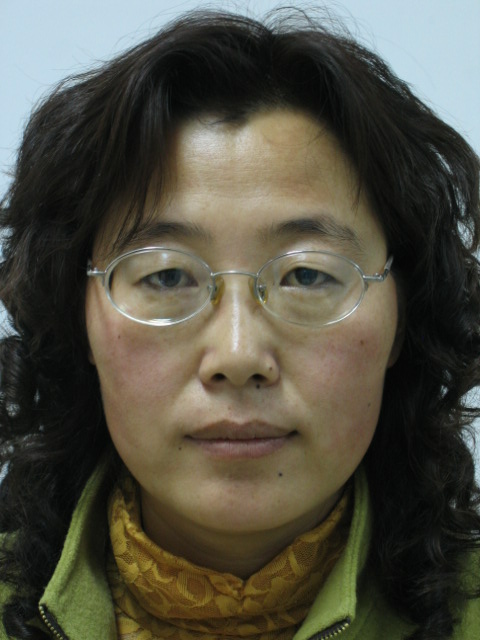} &
\includegraphics[angle=0, width=0.105\textwidth]{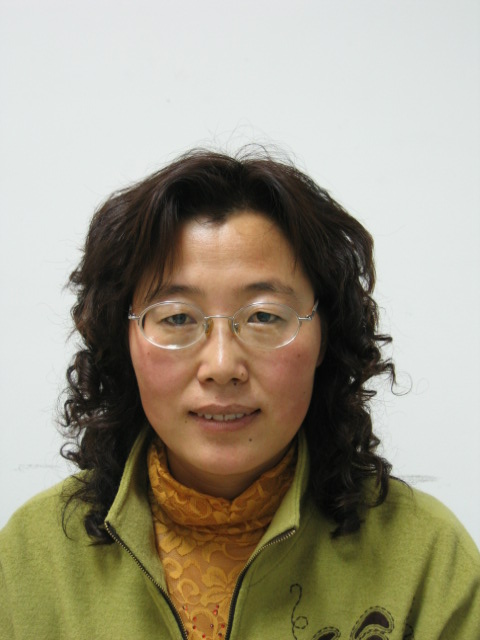} \\
\end{tabular} 
\caption{Sample images of two subjects from the CASIA NIR-VIS 2.0 Face Dataset. Top:
  NIR. Bottom: VIS.}
\label{fig:NIR-VIS-example}
\end{figure}
\begin{table}[b]
\begin{center}
 \small
 \begin{tabular}{|c|c|c|c|c|}
  \hline 
 Ch. & layers & first and last  & intermediate  & \begin{tabular}{c} skip-\\connections\end{tabular} \\
  \hline \hline
  \multirow{3}{*}{Y} &     &  148x11x11         &  36x11x11         & input       \\
                     & 11  & st. 1, pad 5       & st. 1, pad 5      & to          \\
                     &     & PReLU              & PReLU             & last layer  \\
  \hline  
  \multirow{3}{*}{Cb} &     &  66x3x3        &  32x3x3       &             \\
                      & 7   & st. 1, pad 1       & st. 1, pad 1      & none        \\
                      &     & PReLU              & PReLU             &             \\
  \hline  
  \multirow{3}{*}{Cr} &     &  148x5x5       &  48x5x5       &             \\
                      & 8   & st. 1, pad 2       & st. 1, pad 2      & none        \\
                      &     & PReLU              & PReLU             &             \\
  \hline  
 \end{tabular}
\end{center}
\caption{Architecture of the CNN used for
  cross-spectral hallucination.  The first and last layers have
  deeper filters than the layers in-between, mimicking an
  encoding-decoding scheme.}
\label{tab:cross-spectral-cnn-architecture}
\end{table}

\subsection{Mining for NIR-VIS patches}
\begin{figure} [t]
\centering
\includegraphics[angle=0, width=.49\textwidth]{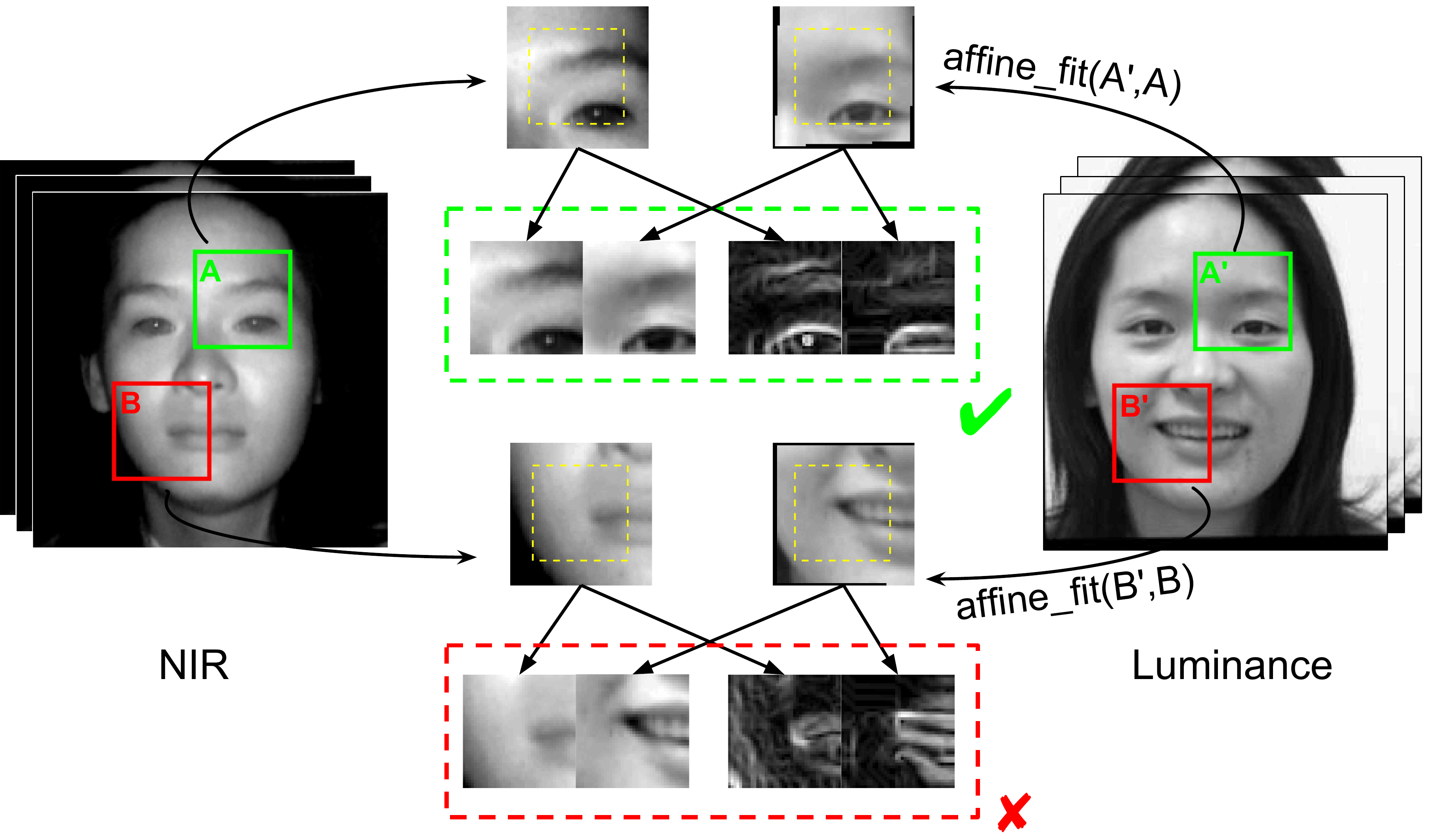}
\caption{Mining the CASIA NIR-VIS 2.0 dataset for valid patch
  correspondences. We compare every NIR image to the luminance channel of every
  VIS image for each subject. Note that the NIR and VIS images are captured
  under different pose and facial expression. We use 224x224 crops of the
  original images with facial landmarks aligned. A sliding 60x60 patch is
  extracted at the same location from both images. The VIS patch is
  affine-registered to the NIR patch. We then crop a 40x40 region inside the
  registered patches. We keep the pair if the correlation of the two patches and
  of their gradients are above a threshold. In this example, patches A
  and A' form a qualifying pair, whilst B and B' are discarded.}
\label{fig:patch_mining}
\end{figure}

We use the CASIA NIR-VIS 2.0 dataset \cite{NIR-VIS-data} to obtain pairs of
NIR-VIS patches. This dataset contains 17,580 NIR-VIS pairs of images with an
average of 24 images of each modality for each subject. This dataset cannot be
used for training directly because the NIR-VIS image pairs are not aligned and
the subjects pose and facial expression vary a lot
(Figure~\ref{fig:NIR-VIS-example}). In \cite{juefei2015nir}, this problem was
partially avoided by subsampling 128x128 crops of the original images to 32x32
images. Yet in their reported results, some smoothing and visual artifacts due
to the training set misalignment can be observed (\cite{juefei2015nir},
Figure~8). In this work we perform no subsampling. Instead, we mine the
CASIA NIR-VIS 2.0 dataset for consistent pairs of NIR-VIS patches at the best
possible resolution, Figure~\ref{fig:patch_mining}. Through this process we
are able to derive a secondary dataset with more than one million pairs of 40x40
NIR-VIS image patches.

We use the luminance channel and the NIR image to find correspondences. The
first step is to pre-process the images by aligning the facial landmarks of the
two modalities (centers of the two eye pupils and center of the mouth), using
\cite{landmark}, and cropping the images to 224x224 pixels. Secondly we
normalize the mean and standard deviation of the NIR and color channels of all
the images in the dataset with respect to a fixed reference chosen from the
training set. The facial landmark alignment is insufficient, as discrepancies
between the NIR and VIS images still occur. Training a CNN with even slightly
inconsistent pairs produces strong artifacts at the output.  In order to obtain
a clean training dataset, we run a sliding window of 60x60 pixels, with stride
12, through both images, and extract patches at the same locations. Note that
the patches are roughly aligned based on the facial landmarks, but this
alignment is typically not fully accurate. We then fit an affine transform
between the 60x60 luminance patch and NIR patch. Next, we crop the center 40x40
regions and compute a similarity score to evaluate if the patches of both
modalities coincide. The similarity score consists of the correlation between
the patches plus the correlation between their gradient magnitudes. If the sum
of both values is above 1 and neither of them is below 0.4 we consider the pair a
valid match. Note that a cross-spectral NIR-VIS patch similarity metric using a
CNN was proposed in \cite{aguilera2016learning}. In this work we opt for using
plain correlation for the sake of efficiency.

\begin{figure} [h!]
  \begin{small}
  \begin{center}
\renewcommand{\arraystretch}{.4}
\begin{tabular}{
    @{\hspace{-1em}}
    c@{\hspace{.2em}}
    c@{\hspace{.2em}}
    c@{\hspace{.2em}}
    c@{\hspace{.2em}}
    c@{\hspace{.2em}}
    |@{\hspace{.2em}}
    c@{\hspace{.2em}}
    c@{\hspace{.2em}}
    c@{\hspace{.2em}}
    c@{\hspace{.2em}}
    c}
NIR & Y & Cb & Cr & RGB & NIR & Y & Cb & Cr & RGB\\
\includegraphics[angle=0, width=.043\textwidth]{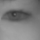} & 
\includegraphics[angle=0, width=.043\textwidth]{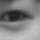} & 
\includegraphics[angle=0, width=.043\textwidth]{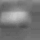} & 
\includegraphics[angle=0, width=.043\textwidth]{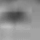} & 
\includegraphics[angle=0, width=.043\textwidth]{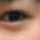} &
\includegraphics[angle=0, width=.043\textwidth]{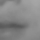} & 
\includegraphics[angle=0, width=.043\textwidth]{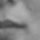} & 
\includegraphics[angle=0, width=.043\textwidth]{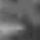} & 
\includegraphics[angle=0, width=.043\textwidth]{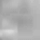} & 
\includegraphics[angle=0, width=.043\textwidth]{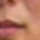} \\
&
\includegraphics[angle=0, width=.043\textwidth]{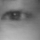} & 
\includegraphics[angle=0, width=.043\textwidth]{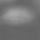} & 
\includegraphics[angle=0, width=.043\textwidth]{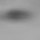} & 
\includegraphics[angle=0, width=.043\textwidth]{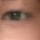} &
&
\includegraphics[angle=0, width=.043\textwidth]{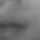} & 
\includegraphics[angle=0, width=.043\textwidth]{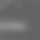} & 
\includegraphics[angle=0, width=.043\textwidth]{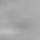} & 
\includegraphics[angle=0, width=.043\textwidth]{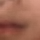} \\
\includegraphics[angle=0, width=.043\textwidth]{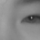} & 
\includegraphics[angle=0, width=.043\textwidth]{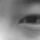} & 
\includegraphics[angle=0, width=.043\textwidth]{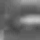} & 
\includegraphics[angle=0, width=.043\textwidth]{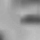} & 
\includegraphics[angle=0, width=.043\textwidth]{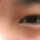} &
\includegraphics[angle=0, width=.043\textwidth]{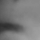} & 
\includegraphics[angle=0, width=.043\textwidth]{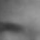} & 
\includegraphics[angle=0, width=.043\textwidth]{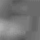} & 
\includegraphics[angle=0, width=.043\textwidth]{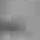} & 
\includegraphics[angle=0, width=.043\textwidth]{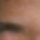} \\
 & 
\includegraphics[angle=0, width=.043\textwidth]{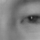} & 
\includegraphics[angle=0, width=.043\textwidth]{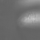} & 
\includegraphics[angle=0, width=.043\textwidth]{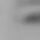} & 
\includegraphics[angle=0, width=.043\textwidth]{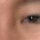} &
 & 
\includegraphics[angle=0, width=.043\textwidth]{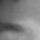} & 
\includegraphics[angle=0, width=.043\textwidth]{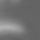} & 
\includegraphics[angle=0, width=.043\textwidth]{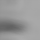} & 
\includegraphics[angle=0, width=.043\textwidth]{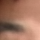} \\
\includegraphics[angle=0, width=.043\textwidth]{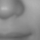} & 
\includegraphics[angle=0, width=.043\textwidth]{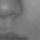} & 
\includegraphics[angle=0, width=.043\textwidth]{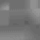} & 
\includegraphics[angle=0, width=.043\textwidth]{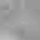} & 
\includegraphics[angle=0, width=.043\textwidth]{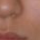} &
\includegraphics[angle=0, width=.043\textwidth]{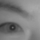} & 
\includegraphics[angle=0, width=.043\textwidth]{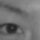} & 
\includegraphics[angle=0, width=.043\textwidth]{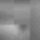} & 
\includegraphics[angle=0, width=.043\textwidth]{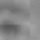} & 
\includegraphics[angle=0, width=.043\textwidth]{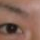} \\
 & 
\includegraphics[angle=0, width=.043\textwidth]{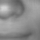} & 
\includegraphics[angle=0, width=.043\textwidth]{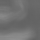} & 
\includegraphics[angle=0, width=.043\textwidth]{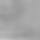} & 
\includegraphics[angle=0, width=.043\textwidth]{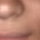} &
 & 
\includegraphics[angle=0, width=.043\textwidth]{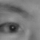} & 
\includegraphics[angle=0, width=.043\textwidth]{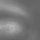} & 
\includegraphics[angle=0, width=.043\textwidth]{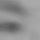} & 
\includegraphics[angle=0, width=.043\textwidth]{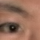} \\
\includegraphics[angle=0, width=.043\textwidth]{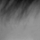} & 
\includegraphics[angle=0, width=.043\textwidth]{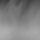} & 
\includegraphics[angle=0, width=.043\textwidth]{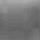} & 
\includegraphics[angle=0, width=.043\textwidth]{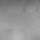} & 
\includegraphics[angle=0, width=.043\textwidth]{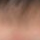} &
\includegraphics[angle=0, width=.043\textwidth]{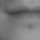} & 
\includegraphics[angle=0, width=.043\textwidth]{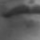} & 
\includegraphics[angle=0, width=.043\textwidth]{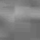} & 
\includegraphics[angle=0, width=.043\textwidth]{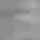} & 
\includegraphics[angle=0, width=.043\textwidth]{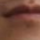} \\
 & 
\includegraphics[angle=0, width=.043\textwidth]{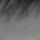} & 
\includegraphics[angle=0, width=.043\textwidth]{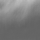} & 
\includegraphics[angle=0, width=.043\textwidth]{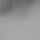} & 
\includegraphics[angle=0, width=.043\textwidth]{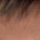} &
 & 
\includegraphics[angle=0, width=.043\textwidth]{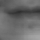} & 
\includegraphics[angle=0, width=.043\textwidth]{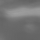} & 
\includegraphics[angle=0, width=.043\textwidth]{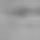} & 
\includegraphics[angle=0, width=.043\textwidth]{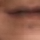} \\
\includegraphics[angle=0, width=.043\textwidth]{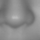} & 
\includegraphics[angle=0, width=.043\textwidth]{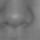} & 
\includegraphics[angle=0, width=.043\textwidth]{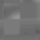} & 
\includegraphics[angle=0, width=.043\textwidth]{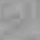} & 
\includegraphics[angle=0, width=.043\textwidth]{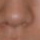} &
\includegraphics[angle=0, width=.043\textwidth]{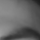} & 
\includegraphics[angle=0, width=.043\textwidth]{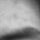} & 
\includegraphics[angle=0, width=.043\textwidth]{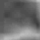} & 
\includegraphics[angle=0, width=.043\textwidth]{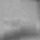} & 
\includegraphics[angle=0, width=.043\textwidth]{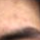} \\
 & 
\includegraphics[angle=0, width=.043\textwidth]{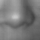} & 
\includegraphics[angle=0, width=.043\textwidth]{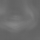} & 
\includegraphics[angle=0, width=.043\textwidth]{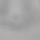} & 
\includegraphics[angle=0, width=.043\textwidth]{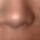} &
 & 
\includegraphics[angle=0, width=.043\textwidth]{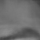} & 
\includegraphics[angle=0, width=.043\textwidth]{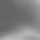} & 
\includegraphics[angle=0, width=.043\textwidth]{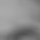} & 
\includegraphics[angle=0, width=.043\textwidth]{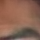} \\
\includegraphics[angle=0, width=.043\textwidth]{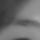} & 
\includegraphics[angle=0, width=.043\textwidth]{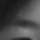} & 
\includegraphics[angle=0, width=.043\textwidth]{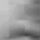} & 
\includegraphics[angle=0, width=.043\textwidth]{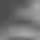} & 
\includegraphics[angle=0, width=.043\textwidth]{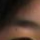} &
\includegraphics[angle=0, width=.043\textwidth]{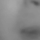} & 
\includegraphics[angle=0, width=.043\textwidth]{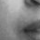} & 
\includegraphics[angle=0, width=.043\textwidth]{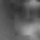} & 
\includegraphics[angle=0, width=.043\textwidth]{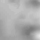} & 
\includegraphics[angle=0, width=.043\textwidth]{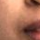} \\
 & 
\includegraphics[angle=0, width=.043\textwidth]{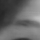} & 
\includegraphics[angle=0, width=.043\textwidth]{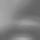} & 
\includegraphics[angle=0, width=.043\textwidth]{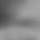} & 
\includegraphics[angle=0, width=.043\textwidth]{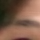} &
&
\includegraphics[angle=0, width=.043\textwidth]{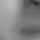} & 
\includegraphics[angle=0, width=.043\textwidth]{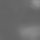} & 
\includegraphics[angle=0, width=.043\textwidth]{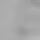} & 
\includegraphics[angle=0, width=.043\textwidth]{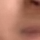} \\
\includegraphics[angle=0, width=.043\textwidth]{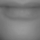} & 
\includegraphics[angle=0, width=.043\textwidth]{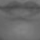} & 
\includegraphics[angle=0, width=.043\textwidth]{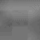} & 
\includegraphics[angle=0, width=.043\textwidth]{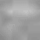} & 
\includegraphics[angle=0, width=.043\textwidth]{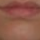} &
\includegraphics[angle=0, width=.043\textwidth]{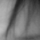} & 
\includegraphics[angle=0, width=.043\textwidth]{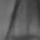} & 
\includegraphics[angle=0, width=.043\textwidth]{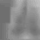} & 
\includegraphics[angle=0, width=.043\textwidth]{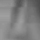} & 
\includegraphics[angle=0, width=.043\textwidth]{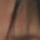} \\
 & 
\includegraphics[angle=0, width=.043\textwidth]{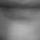} & 
\includegraphics[angle=0, width=.043\textwidth]{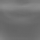} & 
\includegraphics[angle=0, width=.043\textwidth]{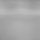} & 
\includegraphics[angle=0, width=.043\textwidth]{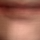} &
 & %
\includegraphics[angle=0, width=.043\textwidth]{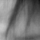} & 
\includegraphics[angle=0, width=.043\textwidth]{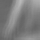} & 
\includegraphics[angle=0, width=.043\textwidth]{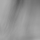} & 
\includegraphics[angle=0, width=.043\textwidth]{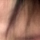} \\
  \end{tabular}
  \end{center}
\caption{Example patches extracted from CASIA-NIS-VIR 2.0 database using the
  proposed patch mining method. For each patch, the top row shows the NIR input
  and ground truth Y, Cb, Cr and RGB signals, and the bottom row shows the
  output of the cross-spectral hallucination CNN. The Cb and Cr values have
  been scaled for better visualization. In total, we were able to mine 1.2
  million such pairs of NIR-VIS patches, equally distributed along the face.  All
  the patches in this figure belong to the validation set and have not been seen
  during training. (Best viewed in electronic format.)}
\label{fig:example_patches}
\end{small}
\end{figure}

This patch mining strategy allowed us to collect more than 700,000 pairs of
NIR-VIS patches. We then pruned this dataset to ensure that the patches were
approximately uniformly distributed around the face. After this pruning we kept
a total of 600,000 patches. We horizontally flip each patch to form a final
dataset of 1.2 million aligned NIR-VIS patches that we use for training and
validation of the cross-spectral hallucination CNN.
Figure~\ref{fig:example_patches} shows examples of obtained input and output
patches, including the result of the hallucination CNN for those patches (not
seen during training). The filters learned by the three networks can be applied
to any NIR image to produce a VIS image equivalent. Note that we retained the
subject identification for each patch so that the dataset can be split without
subject overlap.

\subsection{Post-processing}
\label{sec:blend}

Ideally, one would not like to lose all the rich information contained in the
original NIR image. Despite our methodology for mining aligned patches, it is
not at all impossible that the CNN introduces small artifacts in unseen
patches. To safeguard the valuable details of the original NIR, we propose to
blend the CNN output with the original NIR image. A successful blending smooths
the result of the cross-spectral hallucination and maintains valid information
from the pure NIR image. We will later analyze this fact in the experimental
section. We perform the blending only on the luminance channel, by computing the
image
\begin{equation} \label{eq:blend}
Y = \hat{Y}-\alpha\cdot G_{\sigma}^2*(N_{ir}-\hat{Y}),
\end{equation}
where $Y$ is the final luminance channel estimation, $\hat{Y}$ is the
output of the cross-spectral CNN, $N_{ir}$ is the NIR image,
$G_\sigma$ is a Gaussian filter with $\sigma=1$, and $*$ denotes
convolution. The parameter $\alpha$ balances the amount of
information retained from the NIR images and the information obtained
with the CNN and allows to remove some of the artifacts introduced by
the CNN ($\alpha=0.6$ in our experiments). 

\begin{figure} [t]
  \begin{center}
\renewcommand{\arraystretch}{0.0}
\begin{tabular}{c@{}c@{}c@{}c}
\includegraphics[angle=0, width=.113\textwidth]{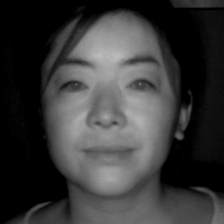} & 
\includegraphics[angle=0, width=.113\textwidth]{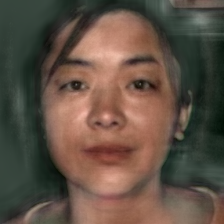} & 
\includegraphics[angle=0, width=.113\textwidth]{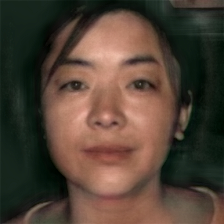} & 
\includegraphics[angle=0, width=.113\textwidth]{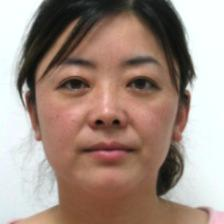} \\ 
\includegraphics[angle=0, width=.113\textwidth]{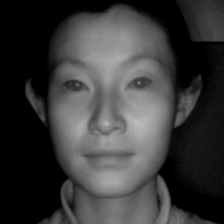} & 
\includegraphics[angle=0, width=.113\textwidth]{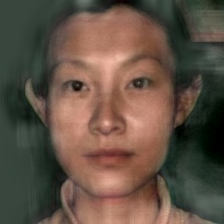} & 
\includegraphics[angle=0, width=.113\textwidth]{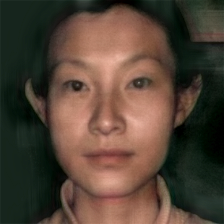} & 
\includegraphics[angle=0, width=.113\textwidth]{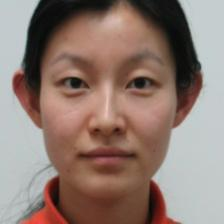} \\ 
\includegraphics[angle=0, width=.113\textwidth]{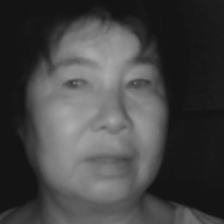} & 
\includegraphics[angle=0, width=.113\textwidth]{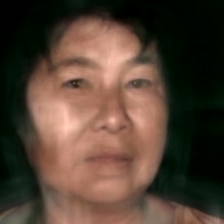} & 
\includegraphics[angle=0, width=.113\textwidth]{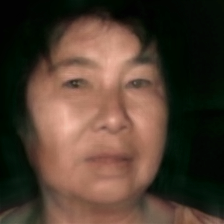} & 
\includegraphics[angle=0, width=.113\textwidth]{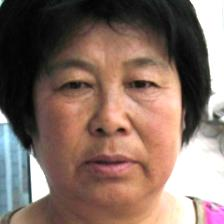} \\ 
\includegraphics[angle=0, width=.113\textwidth]{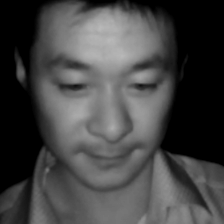} & 
\includegraphics[angle=0, width=.113\textwidth]{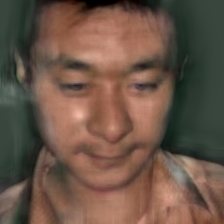} & 
\includegraphics[angle=0, width=.113\textwidth]{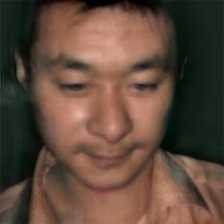} & 
\includegraphics[angle=0, width=.113\textwidth]{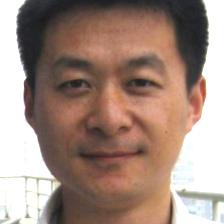} \\ 
\includegraphics[angle=0, width=.113\textwidth]{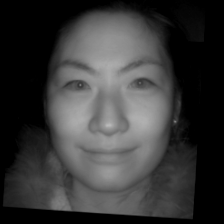} & 
\includegraphics[angle=0, width=.113\textwidth]{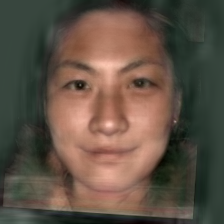} & 
\includegraphics[angle=0, width=.113\textwidth]{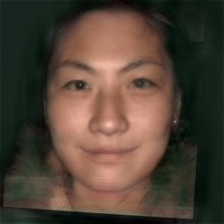} & 
\includegraphics[angle=0, width=.113\textwidth]{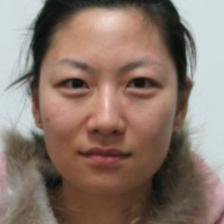} \\ 
\includegraphics[angle=0, width=.113\textwidth]{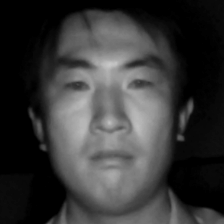} & 
\includegraphics[angle=0, width=.113\textwidth]{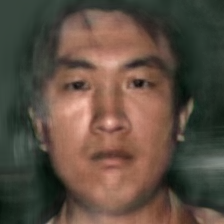} & 
\includegraphics[angle=0, width=.113\textwidth]{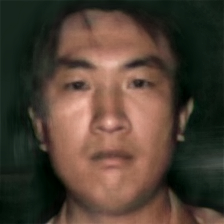} & 
\includegraphics[angle=0, width=.113\textwidth]{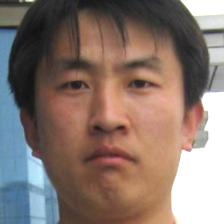} \\ 
\includegraphics[angle=0, width=.113\textwidth]{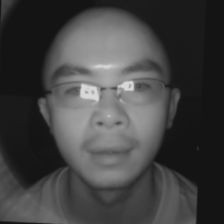} & 
\includegraphics[angle=0, width=.113\textwidth]{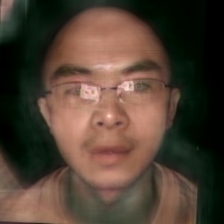} & 
\includegraphics[angle=0, width=.113\textwidth]{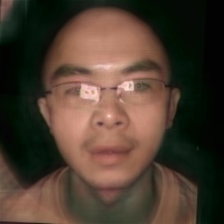} & 
\includegraphics[angle=0, width=.113\textwidth]{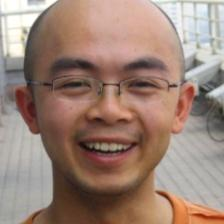} \\ 
\includegraphics[angle=0, width=.113\textwidth]{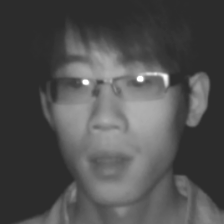} & 
\includegraphics[angle=0, width=.113\textwidth]{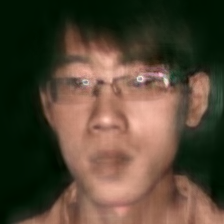} & 
\includegraphics[angle=0, width=.113\textwidth]{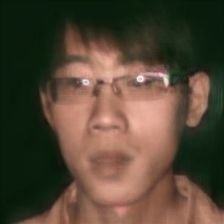} & 
\includegraphics[angle=0, width=.113\textwidth]{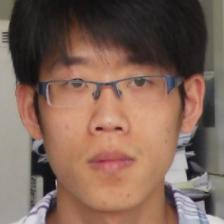} \\ 
\includegraphics[angle=0, width=.113\textwidth]{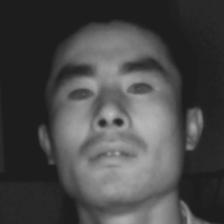} &
\includegraphics[angle=0, width=.113\textwidth]{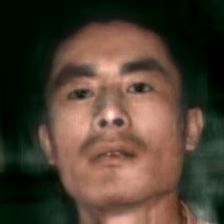} &
\includegraphics[angle=0, width=.113\textwidth]{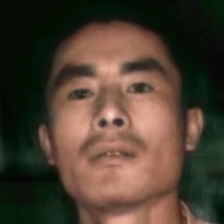} &
\includegraphics[angle=0, width=.113\textwidth]{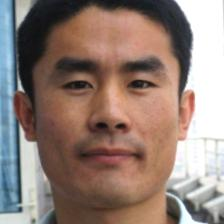} \\
\end{tabular}
\end{center} 
\caption{Results of the deep cross-modal hallucination for subjects in the
  validation set. From left to
  right: Input NIR image; Raw output of the hallucination CNN; output of the CNN
  after post-processing; one RGB sample for each subject. The post-processing
  helps removing some of the artifacts in the CNN output. See for example the
  faces with glasses, which cause the CNN to create notorious artifacts. Note
  that the CNN was trained only on face patches so the color of the clothes
  cannot be hallucinated. (Best viewed in electronic format.)}
\label{fig:colorization_results}
\end{figure}

Figure~\ref{fig:colorization_results} shows example results of the
cross-modal hallucination for subjects not seen during
training. 
Note how the blending helps
correcting some of the remaining artifacts in the CNN output but
maintaining a more natural-looking face than the NIR alone.

\begin{figure*} [t!]
\centering
\subfloat[No embedding]       {\label{fig:no_emb}      \includegraphics[angle=0, height=.19\textwidth, width=.247\textwidth]{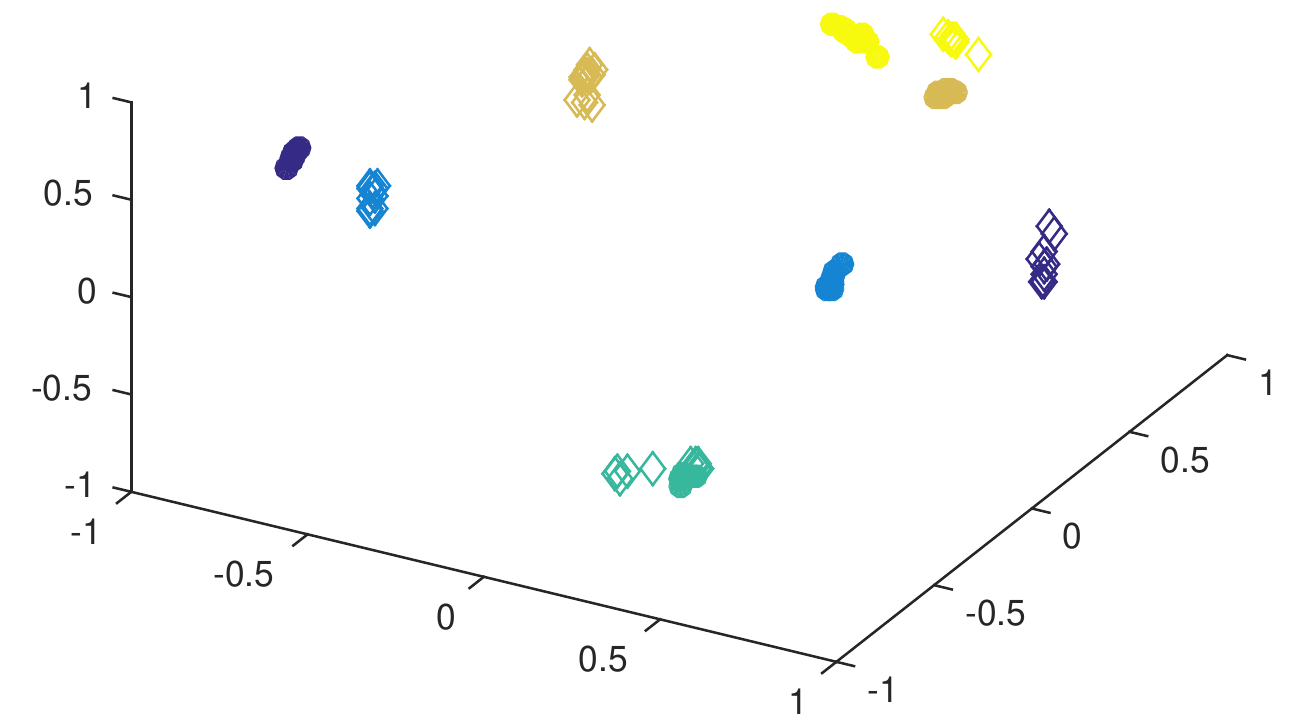}}
\subfloat[Pairwise embedding] {\label{fig:pair_emb}    \includegraphics[angle=0, height=.19\textwidth, width=.247\textwidth]{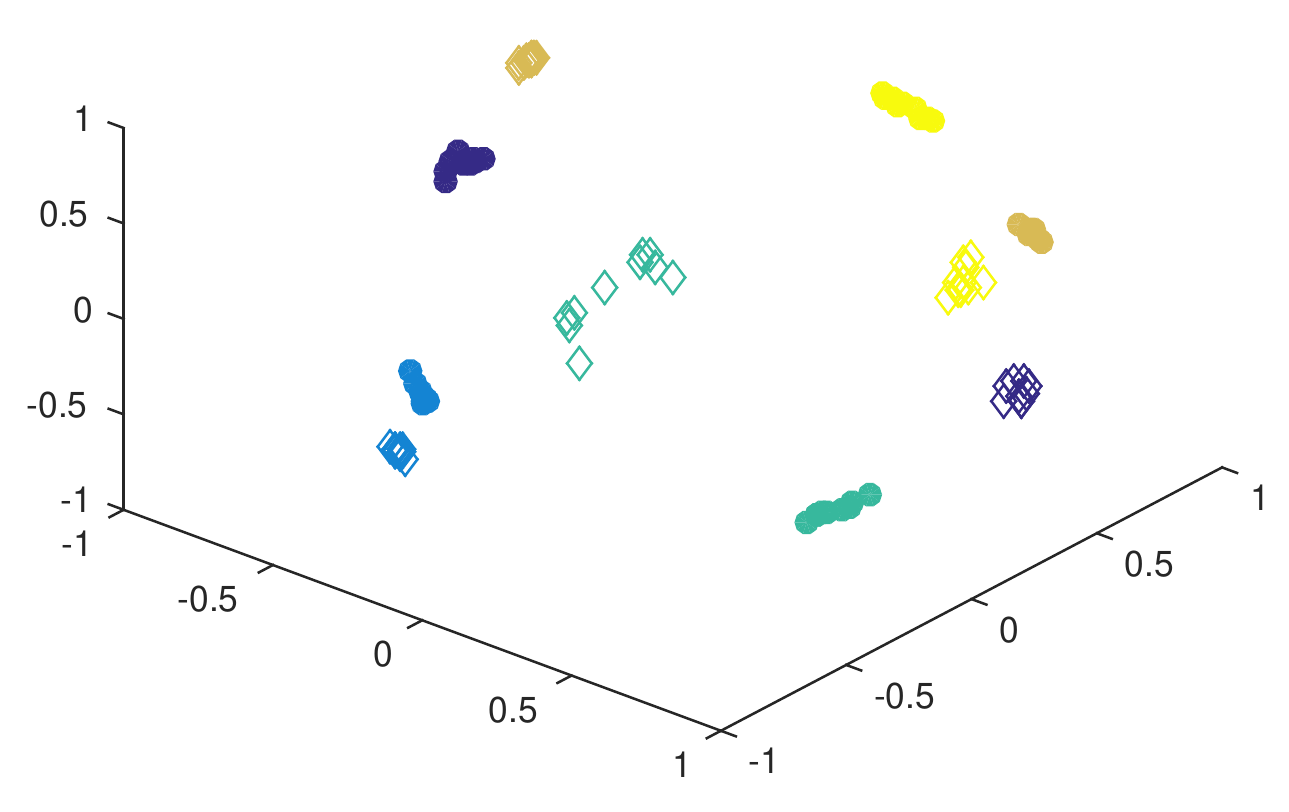}} 
\subfloat[Triplet embedding]  {\label{fig:triplet_emb} \includegraphics[angle=0, height=.19\textwidth, width=.247\textwidth]{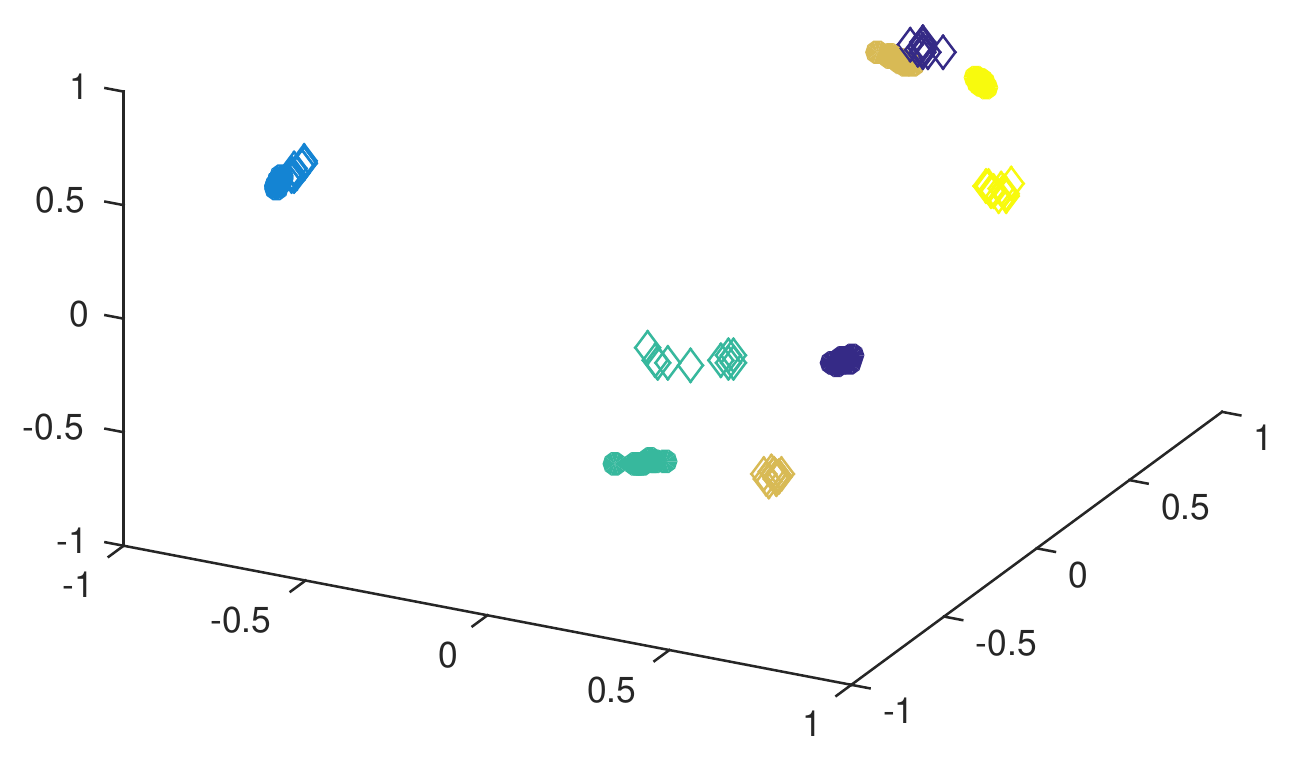}}
\subfloat[Low-rank embedding] {\label{fig:lrt_emb}     \includegraphics[angle=0, height=.19\textwidth, width=.247\textwidth]{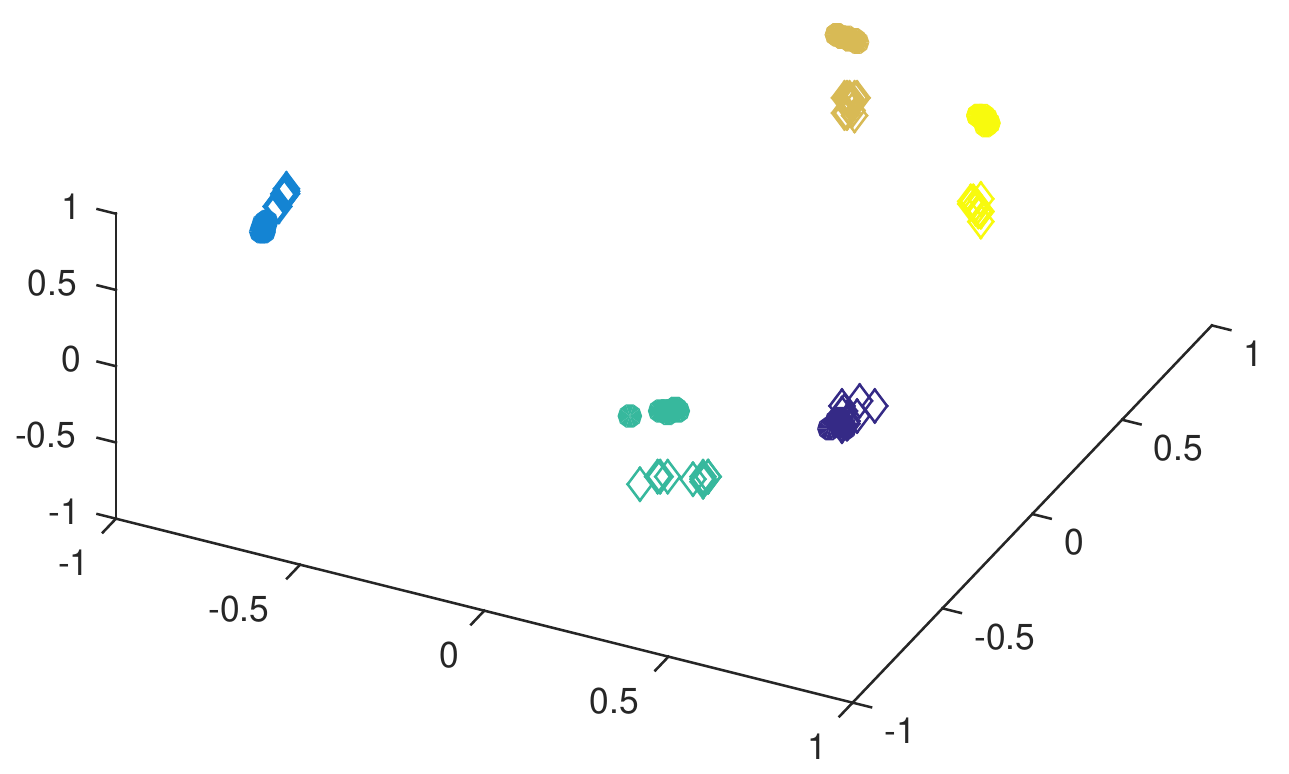}}
\caption{
Deep features generated using the VGG-face model \cite{vggface}  for VIS (filled circle) and NIR (unfilled diamond) face images from five subjects, one color per subject.  Data are visualized in two dimensions using PCA.
In (a), without embedding, 
VIS and NIR faces from the same subject often form two different clusters respectively.
In (d),  low-rank embedding successfully restores a low-rank structure for multi-spectrum faces from the same subject. 
In (b) and (c), popular pair-wise  and triplet embeddings still show significant intra-class variations across spectrums
(best viewed zooming on screen).}
\label{fig:embed}
\end{figure*}

\section{Low-rank Embedding}

In this section, we propose a simple way to extend a DNN model pre-trained on
VIS face images to the NIR spectrum, using low-rank embedding at the output
layer.  The mathematical framework behind low-rank embedding is introduced in
\cite{lrt}, where a geometrically motivated transformation is learned to restore
a within-class low-rank structure, and meanwhile introduce a maximally
separated inter-class structure.  With a low-rank embedding layer appended at
the end, a DNN model that sees only VIS images produces deep features for VIS
and NIR images (or the previously described hallucinated images) in a common
space.

\subsection{Low-rank Transform}

Many high-dimensional data often reside near single or multiple subspaces (or after some non-linear transforms).
Consider a matrix $\mathbf{Y} = \{ \mathbf{y}_i\}_{i=1}^N \subseteq \mathbb{R}^d$, where each column $\mathbf{y}_i$ is a data point in one of the $C$ classes. Let $\mathbf{Y}_c$ denote the submatrix formed by columns of $\mathbf{Y}$ that lie in the $c$-th class. 
A $d \times d$ low-rank transform $\mathbf{T}$ is learned to minimize
\begin{align} \label{nuclear_obj}
{ \sum_{c=1}^C ||\mathbf{T Y}_c||_* - ||\mathbf{T Y}||_*,}
\end{align}
where  $||\mathbf{\cdot}||_*$ denotes the matrix nuclear norm, i.e., the sum of the singular values of a matrix. 
The nuclear norm is the convex envelope of the rank function over the unit ball of matrices \cite{rank-min}. 
An additional condition  $||\mathbf{T}||_2 = 1$ is originally adopted to prevent the trivial solution $\mathbf{T}=0$. 
We drop this normalization condition in this paper, as we never empirically observe such trivial solution, with $\mathbf{T}$ being initialized as the identity matrix.
The objective function (\ref{nuclear_obj}) is a difference of convex functions program,
and the minimization is guaranteed to converge to a local minimum (or a stationary point) using the concave-convex procedure \cite{dc2, dc1}.

\begin{theorem} \label{nuclear_ineq}
Let $\mathbf{A}$ and $\mathbf{B}$ be matrices of the same row dimensions, and $\mathbf{[A,B]}$ denote their column-wise concatenation. Then,
$
||\mathbf{[\mathbf{A},\mathbf{B}]}||_*  \le ||\mathbf{A}||_* + ||\mathbf{B}||_*,
$
 and equality holds if the column spaces of $\mathbf{A}$ and $\mathbf{B}$ are orthogonal.
\end{theorem}
\noindent \emph{Proof.}  This results from properties of the matrix nuclear norm.

Based on Theorem~\ref{nuclear_ineq}, the objective function (\ref{nuclear_obj}) is non-negative, and it achieves the value zero if, after applying the learned transformation $\mathbf{T}$, the column spaces corresponding to different classes become orthogonal (that is, the smallest principal angle between two different subspaces is $\frac{\pi}{2}$). Note that minimizing each of the nuclear norms appearing in (\ref{nuclear_obj})  serves to reduce the variation within a class.
Thus, the low-rank transform simultaneously minimize intra-class variation and maximize inter-class separations.

\subsection{Cross-spectral Embedding}

In traditional single-spectrum VIS face recognition, DNN models adopt the classification objective such as softmax at the final layer.
Deep features produced at the second-to-last layer are often $l_2$-normalized, and then compared using the cosine similarity to perform face recognition \cite{vggface, deepID2}.
Thus, successful face DNN models expect deep features generated for VIS faces from the same subject to reside in a low-dimensional subspace.

In Figure~\ref{fig:no_emb} we illustrate the following, which motivates the
proposed embedding: Face images from five subjects in VIS and NIR are input to
VGG-face \cite{vggface}, one of the best publicly available DNN face models.
The generated deep features are visualized in two dimensions using PCA, with
filled circle for VIS, unfilled diamond for NIR, and one color per subject.  We
observe that VIS and NIR faces from the same subject often form two different
clusters respectively.  Such observation indicates that a successful DNN face
model pre-trained on VIS faces is able to generate discriminative features for
NIR faces; however, when a subject is imaged under a different spectrum, the
underlying low-rank structure assumption is often violated.

Our finding is that the low-rank transform $\mathbf{T}$ in (\ref{nuclear_obj})
can still effectively restore for the same subject a low-rank structure, even
when $\mathbf{Y}_c$ contains mixed NIR and VIS training data from the c-th
subject.  As no DNN retraining is required, a very important advantage of our
approach in practice, the learned low-rank transform can be simply appended as a
linear embedding layer after the DNN output layer to allow a VIS model accepting
both VIS and NIR images.  As shown in Figure~\ref{fig:lrt_emb}, low-rank
embedding effectively unifies cross-spectral deep features in
Figure~\ref{fig:no_emb}.

In DNN-based face recognition, deep feature embeddings with pairwise or triplet
constraints are commonly used \cite{vggface, saxena2016heterogeneous,
  schroff2015facenet}.  Two popular DNN embedding schemes, pair-wise (ITML
\cite{itml}) and triplet (LMNN \cite{lmnn}) embeddings, are shown in
Figure~\ref{fig:pair_emb} and Figure~\ref{fig:triplet_emb} respectively, and
contrary to our approach, significant intra-class variations, i.e., the distance
between same color clusters, are still observed across spectrums.

\section{Experimental Evaluation}

We consider pre-trained VIS DNNs as black-boxes, thereby enjoying
the advances in VIS recognition, and perform cross-spectral hallucination to
input NIR images, and/or low-rank embedding to output features, for
cross-spectral face recognition.  To demonstrate that our approach is generally
applicable for single-spectrum DNNs without any re-training, we experiment with
three pre-trained VIS DNN models from different categories:

\begin{itemize*}
\item The \textbf{VGG-S} model is our trained-from-scratch DNN face model using the VGG-S architecture in \cite{vggs}.

\item  The \textbf{VGG-face} model is a publicly available DNN face model,\footnote{Downloaded at \url{http://www.robots.ox.ac.uk/~vgg/software/vgg_face}.} 
which reports the best result on the LFW face benchmark among publicly available face models.

\item The \textbf{COTS} model is a commercial off the shelf (COTS) DNN face model to which we have access.
\end{itemize*}

\subsection{Dataset}

The CASIA NIR-VIS 2.0 Face Dataset \cite{NIR-VIS-data} is used to evaluate NIR-VIS face recognition
performance. This is the largest available  NIR-VIS face recognition dataset and
contains 17,580 NIR and VIS face images of 725 subjects. This dataset
presents variations in pose, illumination, expressions, etc. 
Sample images are shown from this dataset in Figure~\ref{fig:NIR-VIS-example}.  
The CASIA-Webface dataset \cite{yi2014learning} is used
 to train our trained-from-scratch VGG-S model. CASIA-Webface is one of the
largest public face datasets containing 494,414 VIS face images from
10,575 subjects collected from IMDB.

\subsection{Hallucination Networks Protocol}
We first train the three CNNs used to hallucinate VIS faces from input NIR
images described in Table~\ref{tab:cross-spectral-cnn-architecture}. We use our
mined dataset of NIR-VIS image patches. Given that not all the images in the
CASIA NIR-VIS~2.0 dataset provide the same amount of aligned patches, the
standard protocol for this dataset, which splits the dataset in two, does not
provide enough training data for the cross-spectral hallucination CNN. For that
reason, to properly evaluate the hallucination contribution we split the dataset
into 6 folds. We use five folds (1,030,758 pairs of patches) for training and
one fold (206,151 pairs) for testing.\footnote{Our data partition protocol is
  included in supplementary material, and will be available to the public for
  reproducing our experimental results.} The folds are not arbitrary, but follow
the natural order of the numbering scheme of the original dataset. We make sure
that there is no subject overlap between the training and testing dataset.

We implement the luminance and chroma hallucination networks in the Caffe
learning framework. We train the three networks using ADAM optimization
\cite{kingma2014adam}, with initial learning rate $10^{-5}$, and the standard
parameters $\beta_1=0.9$, $\beta_2=0.999$, $\varepsilon=10^{-8}$. We observe it
is enough to train the networks for 10 epochs.


\begin{table}
\centering
{
	\begin{tabular}{l r}
	\hline
 & Accuracy (\%)\\
	\hline
	\hline	
VGG-S&  75.04   \\
VGG-S + Hallucination &  80.65   \\
VGG-S + Low-rank &   89.88 \\
VGG-S + Hallucination + Low-rank &  95.72  \\
	\hline
VGG-face&   72.54 \\
VGG-face + Hallucination &    83.10  \\
VGG-face + Low-rank &  82.26  \\
VGG-face + Hallucination + Low-rank &  91.01  \\
	\hline
COTS&  83.84  \\
COTS + Hallucination &  93.02  \\
COTS + Low-rank &  91.83 \\
COTS + Hallucination + Low-rank &  \textbf{96.41} \\
\hline	
	\end{tabular}
}	

\caption{ Cross-spectral rank-1 identification rate on CASIA NIR-VIS 2.0 (see
  text for protocol).  We evaluate three pre-trained single-spectrum (VIS) DNN
  models: VGG-S, VGG-face, and COTS.  This experiment shows the effectiveness of
  cross-spectrally hallucinating an NIR image input, or low-rank embedding the
  output (universally for all the tested DNNs). When both schemes are used
  together, we observe significant further improvements, e.g., 75.04\% to
  95.72\% for the VGG-S model. The proposed framework gives state-of-the-art
  (96.41\%) without touching at all the VIS recognition system.  }
	\label{tab:rank1_acc_color}
\end{table}

\begin{table}
\centering
{
	\begin{tabular}{l r}
	\hline
 & Accuracy (\%)\\
	\hline
	\hline	
Jin et al. \cite{Jin_etal} &  75.70 $\pm$ 2.50 \\
Juefei-Xu et al. \cite{juefei2015nir} &  78.46 $\pm$ 1.67  \\
Lu et al. \cite{lu_etal} &   81.80 $\pm$ 2.30 \\
Saxena et al. \cite{saxena2016heterogeneous} &  85.90 $\pm$ 0.90 \\
Yi et al. \cite{Yi_etal} &  86.16 $\pm$  0.98\\
Liu et al. \cite{liu2016transferring} &  \textbf{95.74} $\pm$ \textbf{0.52}\\
	\hline
VGG-S&   57.53  $\pm$  2.31   \\
VGG-face&  66.97 $\pm$ 1.62  \\
COTS &  79.29 $\pm$  1.54 \\
	\hline
VGG-S + Triplet &  67.13 $\pm$ 3.01 \\
VGG-face + Triplet &   75.96 $\pm$   2.90    \\
COTS + Triplet &  84.91 $\pm$ 3.32 \\
	\hline	
VGG-S + Low-rank & 82.07  $\pm$  1.27  \\
VGG-face + Low-rank &  80.69  $\pm$  1.02  \\
COTS + Low-rank & \textit{89.59}  $\pm$  \textit{0.89}  \\
\hline
	\end{tabular}
}	

\caption{ Cross-spectral rank-1 identification rate on the 10-fold CASIA NIR-VIS~2.0
  benchmark.  The evaluation protocol defined in \cite{NIR-VIS-data} is adopted.
  We evaluate three single-spectrum DNN models: VGG-S, VGG-face, and COTS.
  Single-spectrum DNNs perform significantly better for cross-spectral
  recognition by applying the proposed low-rank embedding at the output
  (universally for all the tested DNNs), which is a simple linear transform on
  the features.  The popular triplet embedding \cite{lmnn} shows inferior to
  low-rank embedding for such a cross-spectrum task.  Excluding
  \cite{liu2016transferring}, we report the best result on this largest VIS-NIR
  face benchmark. As discussed before, \cite{liu2016transferring} tunes/adapts
  the network to the particular dataset, achieving results slightly below to
  those we report in Table \ref{tab:rank1_acc_color} for our full system;
  results we obtain without any need for re-training and thereby showing the
  generalization power of the approach, enjoying the advantages of both existing
  and potentially new-coming VIS face recognition systems.  }
	\label{tab:rank1_acc}
\end{table}


\subsection{Cross-spectral Face Recognition Protocol}

Our goal is to match a NIR probe face image with VIS face images in the gallery.
We consider three pre-trained single-spectrum DNNs, VGG-S, VGG-face and COTS, as
black-boxes, and only modify their inputs (cross-spectral hallucination)
and/or outputs (low-rank embedding) for cross-spectral face recognition.  All
three models expect RGB inputs. When no VIS hallucination is used, we replicate
the single-channel NIR images from the CASIA NIR-VIS~2.0 dataset into three channels
to ensure compatibility. When hallucination is used, we first apply the
hallucination CNN to the NIR images and then feed the hallucinated VIS images to
the single-spectrum DNN.

We generate deep features from the respective DNN models, which are reduced to
1024 dimensions using PCA.  We then learn a 1024-by-1024 low-rank transform
matrix to align the two spectrums.  Note that, for efficiency, the PCA and
low-rank transform matrices can be merged. We use cosine similarity to perform
the matching.

\subsection{Results}
We evaluate the performance gain introduced by cross-spectral hallucination and
the low-rank transform, and by both techniques combined.  As explained, our
cross-spectral hallucination CNN requires more training data than the one
available in the standard CASIA NIR-VIS~2.0 benchmark training set, so we define a
new splitting of that dataset into 6 folds. We use the same protocol for VIS
hallucination as for face recognition, i.e. five folds for training and one fold
for testing (our data split protocol will be released to public).  There is no
subject overlap between our training and testing partitions.  The evaluation is
performed on the testing partition, using VIS images as the gallery and NIR
images as the probe.

In Table~\ref{tab:rank1_acc_color} we report the rank-1 performance score of the
single-spectrum DNN, with and without hallucination and with and without the
low-rank embedding for the three face models. The results show that it is often
equally effective to hallucinate a VIS image from the NIR input, or to low-rank
embed the output. Both schemes independently introduce a significant gain in
performance with respect to using single-spectrum DNNs.  When hallucination and
low-rank embedding are used together, we observe significant further
improvements, e.g., from 75.04\% to 95.72\% for VGG-S. Using the COTS and the
combination of hallucination and low-rank embedding, the proposed framework
yields a state-of-the-art 96.41\% rank-1 accuracy, without touching at all the
VIS pre-trained DNN and with virtually no additional computational cost.

For completeness, we also present our results using the low-rank transform for
the standard CASIA NIR-VIS~2.0 evaluation protocol in Table~\ref{tab:rank1_acc}
(recall that the standard protocol is not possible for the added hallucination
step). These results demonstrate that the low-rank embedding dramatically
improves single-spectrum DNNs VIS-NIR rank-1 identification rate, 57.53\% to
82.07\% for VGG-S, 66.97\% to 80.69\% for VGG-face, and 79.29\% to 89.59\% for
COTS. One of the most effective triplet embedding methods, LMNN \cite{lmnn},
shows inferior performance to the proposed low-rank embedding for this
cross-spectrum task.

The results obtained by our full system (Table \ref{tab:rank1_acc_color}) and
our partial system (Table~\ref{tab:rank1_acc}), indicate that the combination of
hallucination and low-rank embedding produce state-of-the-art results on
cross-spectral face recognition, without having to adapt or fine-tune an
existing deep VIS model.

\begin{figure} [ht]
\centering
\includegraphics[angle=0, width=0.33\textwidth]{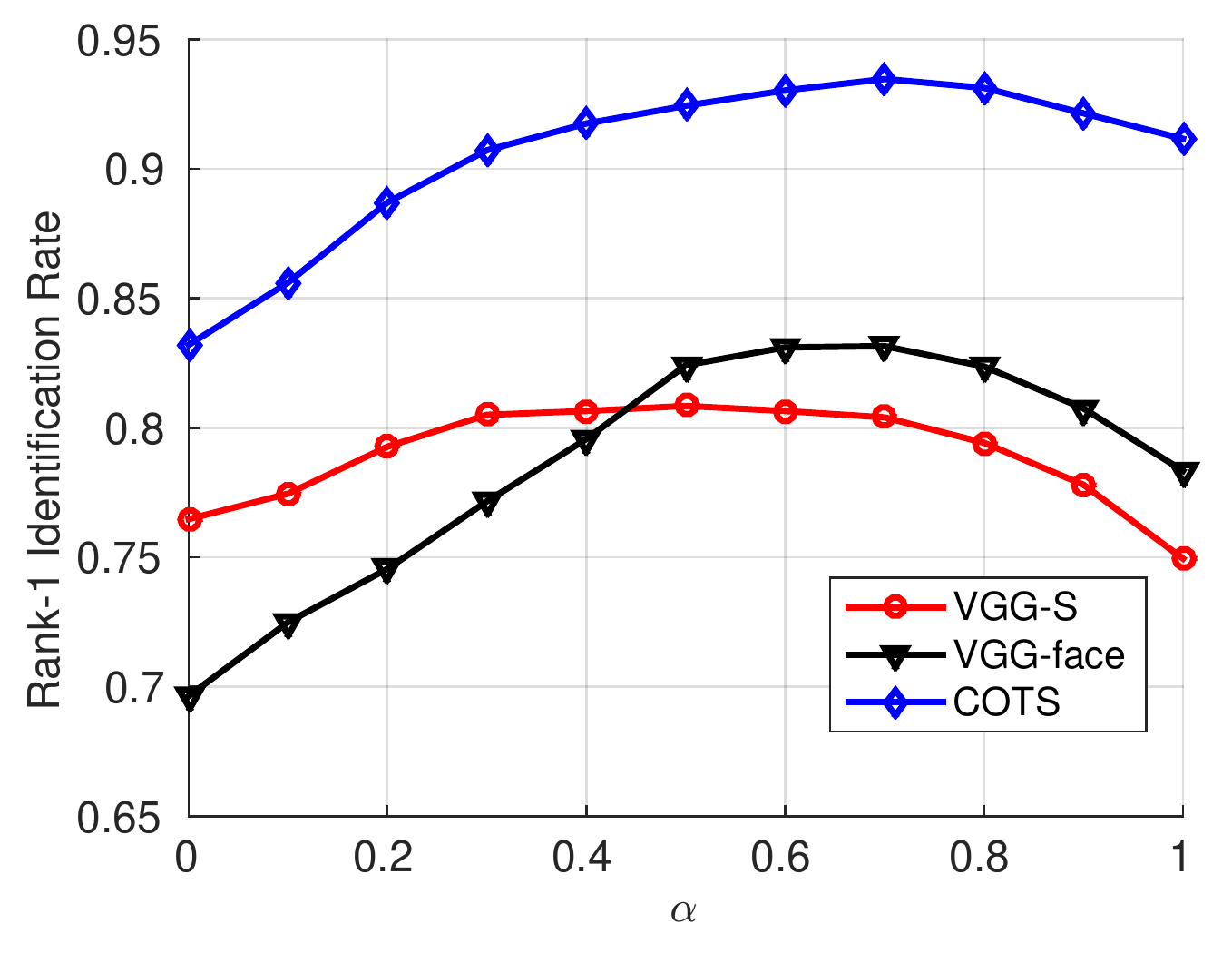}
\caption{The impact of the blending parameter $\alpha$ in (\ref{eq:blend}) on face recognition. We evaluate three single-spectrum DNN models: VGG-S, VGG-face, and COTS.}
\label{fig:blend_impact}
\end{figure}

As discussed in Section~\ref{sec:blend}, to preserve the details of the original
NIR, we blend the hallucinated luminance channel of the CNN output with the
original NIR image to remove possible artifacts introduced by the CNN.
Figure~\ref{fig:blend_impact} shows the impact of the blending parameter
$\alpha$ in (\ref{eq:blend}) on face recognition.  The parameter $\alpha \in [0,
  1]$ balances the amount of information retained from the NIR images and the
information obtained with the hallucination CNN. We usually observe the peak
recognition performance when $\alpha$ is around 0.6-0.7. With $\alpha=0.6$ we
also obtain a more natural-looking face; this is the value used in
Table~\ref{tab:rank1_acc_color} and Figure~\ref{fig:colorization_results}.

\section{Conclusion}
\label{sec:conclusion}

We proposed an approach to adapt a pre-trained state-of-the-art DNN, which has
seen only VIS face images, to generate discriminative features for both VIS and
NIR face images, without retraining the DNN.  Our approach consists of two core
components, cross-spectral hallucination and low-rank embedding, to adapt DNN
inputs and outputs respectively for cross-spectral recognition.  Cross-spectral
hallucination preprocesses the NIR image using a CNN that performs a
cross-spectral conversion of the NIR image into the VIS spectrum.  Low-rank
embedding restores a low-rank structure for cross-spectral features from the
same subject, while enforcing a maximally separated structure for different
subjects.  We observe significant improvement in cross-spectral face recognition
with the proposed approach.  This new approach can be considered a new direction
in the intersection of transfer learning and joint embedding.

\section*{Acknowledgments}

Work partially supported by ONR, NGA, ARO, NSF, ANII (Uruguayan agency) Grant PD\_NAC\_2015\_1\_108550.

{\small
\bibliographystyle{ieee}
\bibliography{nir}
}

\end{document}